# Written and spoken corpus of real and fake social media postings about COVID-19

Ng Bee Chin, Ng Zhi Ee Nicole, Kyla Kwan, Lee Yong Han Dylann, Liu Fang, Xu Hong

## Abstract

This study investigates the linguistic traits of fake news and real news. There are two parts to this study: text data and speech data. The text data for this study consisted of 6420 COVID-19 related tweets re-filtered from Patwa et al. (2021). After cleaning, the dataset contained 3049 tweets, with 2161 labeled as 'real' and 888 as 'fake'. The speech data for this study was collected from TikTok, focusing on COVID-19 related videos. Research assistants fact-checked each video's content using credible sources and labeled them as 'Real', 'Fake', or 'Questionable', resulting in a dataset of 91 real entries and 109 fake entries from 200 TikTok videos with a total word count of 53,710 words. The data was analysed using the Linguistic Inquiry and Word Count (LIWC) software to detect patterns in linguistic data. The results indicate a set of linguistic features that distinguish fake news from real news in both written and speech data. This offers valuable insights into the role of language in shaping trust, social media interactions, and the propagation of fake news.

## 1 Introduction

This study looks at how specific linguistics features can be used to distinguish real news from fake news. The rise of online media and the instantaneous nature of information dissemination have resulted in a decline in traditional journalism's credibility and reliability as a source of public information (Lee, 2019). This issue has encouraged research into the development of fake news detection technologies, specifically through the analysis of 'style' as defined by Zhou & Zafarani (2018). Machine learning models have shown promise in differentiating fake news using various linguistic measures (Choudhary & Arora, 2021; Santos et al., 2020). However, theoretical explanations for the effectiveness of specific stylistic features and cross-study agreement remain limited.

## 2    Hypotheses and research aims

The overall aim of the study is to investigate and compare the linguistic characteristics of fake news and real news, specifically focusing on perceptual and cognitive process words. The study also aims to explore various linguistic features in both types of news, such as sentence lengths, emotion words, sensationalism, readability, and informality. By analyzing these linguistic aspects, the study seeks to understand potential differences between fake news and real news and provide insights into how language is used in the context of trust, social media, and misinformation.

## 3    Methodology

This study utilises two datasets related to the COVID-19 pandemic: text data via Twitter posts and speech data via TikTok posts. Each dataset was analysed separately in LIWC, run using all LIWC categories based on the LIWC-22 dictionary.

### 3.1 Text data

The text data is a corpus obtained from Patwa et al. (2021), comprising 6420 COVID-19 related tweets. Tweets considered 'real news' are gathered from official Twitter accounts like government agencies and health authorities, while fake news posts are those that were evaluated as fake from factchecking websites such as PolitiFact, Poynter, and Snopes. Upon closer analysis, the dataset was found to be inaccurate. This is especially in the fake tweets sample where at least 2/3 of the tweets were misclassified or contained duplicates. The Patwal et al. (2021) corpus was cleaned through partial automation. First, a team of research assistants manually labelled tweets according to their veracity or category. The dataset was refiltered to remove posts from the fake news sample that could not strictly be considered fake. Table 1 shows the category labels utilized by the research assistants when a tweet could not be strictly categorized as 'fake'. For example, tweets from reputable fact-checking accounts (e.g., PolitiFact) that quoted and called out misinformation were labelled as fake news tweets. Other tweets that contained obviously outlandish claims from satirical accounts were labelled as fake news as well. Such tweets were thus reclassified as 'callouts' and 'satire' respectively and removed from the fake news sample.

*Table 1: List of categories labelled in the social media text posts dataset.*

| Category | Examples |
|---|---|
| **Irrelevant**: Text that is not related to COVID-19 (or link is unclear) | - Elon Musck To New Baby; Get A Job Kid!<br><br>- 10 Road Trip Essentials https://t.co/sB9hzYG2MK #coronavirus #travel #vacation #list #tipsadvice #roadtrips |
| **Non-news**: Text that includes personal opinions, subjective statements, and calls for action. | - MathGuy_7 HedgeyeDJ Thank you - extremely insightful. Is there one or two trusted websites that you follow to provide a good summary & update on covid? I follow chrismartenson and he does a really good job. Recent video touted importance of Vitamin D and elderberry, among other things.<br><br>- DrJudyAMikovits What promises that a viable vaccine can be developed for COVID-19 virus. So, i think they should rehire Judith Mikovits who is a living scientific encyclopedia on viral research. She was involved with the scientific research that permitted<br><br>- We're debunking misinformation about coronavirus. Why? Because hoaxes travel fast and the numbers and causes of this growing outbreak need clarification.<br><br>- I am also doing this, taking Vitamin D for COVID. I think the evidence is moderate-ish right now, not conclusive. But there's no real downside at all to taking moderate doses AFAIK. And people living in the northern latitudes in the winter probably need Vitamin D anyways. |
| **Callout**: Text that serves as a warning about fake news. Includes | - Experts Call Out Claims That Cow Dung/Urine, Yoga, AYUSH Can Prevent Or Treat COVID-19 |
| clarifying an existing piece of fake news, with or without real information added. | - WHO warns against consuming cabbage to prevent COVID-19 |
| **Vague:** For texts where one is unable to tell whether the content is true or false based on text alone. | - Did You Already Have Coronavirus? Here's A Simple Test<br><br>- Where Are They Now: Covid-19<br><br>- Free Coronavirus Test Kits Available To All<br><br>- "Covid is never going away! This is the beach today in Raleigh, North Carolina." |
| **Satire:** Text that is overtly untrue or impossible and likely to be intentionally fatuous, to the point where the average person would not consider the content as fact. | - Orwell not dead but living under a bridge (continued)<br><br>- Covid-19 totally confused about what it's allowed to do where<br><br>- God Has Coronavirus |
| **Real:** Text that contains a claim made related to COVID-19 that can be verified by credible sources. | - With 10576 new infections Maharashtra records highest single-day spike in the state so far. Starting 2pm tomorrow Manipur to go under a complete lockdown for 14 days. Bhopal in Madhya Pradesh to impose a city-wide lockdown from July 24 to August 3.<br><br>- WHO Director General Tedros Adhanom says COVID-19 outbreak is accelerating and we've not yet reached the peak globally. Jharkhand CM Hemant Soren goes into isolation after his colleagues test positive for COVID-19. |

| **Fake:** Text that contains a claim related to COVID-19 that has been factchecked and found to be false. | - BREAKING NEWS The president Cryill Ramaphosa has asked all foreign nations to depart south Africa before 21 june 2020 due to increasing cases of COVID 19 .<br><br>- Finally a INDIAN student from PONDICHERRY university, named RAMU found a home remedy cure for Covid-19 which is for the very first time accepted by WHO. |
|---|---|

Next, the research assistants tagged the text's type and original source (e.g., Twitter, Facebook, WhatsApp). The research assistants also removed any duplicate tweets and non-primary sources from the corpus. The remaining data's formatting and punctuation were then cleaned via automation. This included formatting errors in the dataset, where quotation marks in the original tweets were replaced with question marks in the data. The removal of URLs, hashtag symbols (#), and account mentions symbols (@) from the tweets was also automated.

After this clean-up process, 3049 tweets remained. 2161 of the tweets were labelled as 'real', while the 888 remaining tweets were labelled as 'fake'. Table 2 lists the proportion of text sources.

*Table 2: Number of Fake and Real news posts by text source.*

|  | **Fake** | **Real** | **Total** |
|---|---|---|---|
| Facebook | 544 | 50 | 594 |
| Tweet | 221 | 2099 | 2320 |
| Image | 54 | 0 | 54 |
| Whatsapp | 59 | 0 | 59 |
| Instagram | 10 | 12 | 22 |
| Total | 888 | 2161 | 3049 |

## 3.2 Speech data

The speech data was self-sourced via TikTok, focusing on COVID-19 related videos. A team of research assistants gathered the TikTok videos by searching for COVID-19-related hashtags such as #COVIDVaccine and #COVIDVaccination. This search avoided the use of hashtags with overtly biased connotations such as #COVIDDanger. Educational videos (i.e., those that provided factual information regarding the COVID-19 vaccine) were included in the dataset, while those that were not informational (i.e., for entertainment), were not included. The videos were gathered using a new TikTok account created specifically for this purpose to minimize algorithmic bias.

For each video, metadata such as the TikTok URL, creator's username, number of followers, number of likes, and number of views were also recorded. Each entry was logged into the dataset, with the recording of the text caption accompanying the video, transcribed speech from the video, and on-screen captions as seen within the video. TikTok videos that were found to have no audible speech (having only captions and on-screen subtitles) were also omitted from the analyses.

The research assistants labelled each entry for the video's veracity. Factchecking was done by corroborating each TikTok's content with the researchers' existing knowledge and at least 3 other credible online sources (e.g., government websites, health authorities, and reputable news sites). Videos with verifiable information were labelled 'real', while those that failed to meet the requirements were labelled 'fake'. Videos that were found to express subjective opinions and experiences (e.g., an individual's experience as a COVID-19 patient) were categorised as 'Questionable', as such content would be difficult to strictly classify as true or false. The resulting dataset comprised 91 Real entries and 109 Fake entries, giving 200 TikTok videos and a total word count of 53,710 words. Table 3 shows an example of a transcript from the TikTok videos.

*Table 3: Example of TikTok video transcription.*

| Caption | Speech | On-Screen Text |
|---|---|---|
| Truth about Vaccine! | Fierlafin, I am a medical doctor from Belgium<br><br>specialise in chronic infectious diseases, such as Lyme, et cetera.<br><br>COVID-19 vaccine is not proven safe nor effective<br><br>and I think it is unacceptable that all liabilities have been waived<br><br>for the companies that are producing it. | Do not let the media<br><br>Manipulate you |

| | If Pharma does not take responsibility, for the product they make, | |
| | how can they expect doctors to inject them to the patients | |
| | without doubt of doing harm? | |
| | More and more, we see that this is really not a medical pandemic. | |
| | The measures for Corona cause far more collateral damage | |
| | than the virus caused itself. | |
| | Worldwide, we see that the numbers of cases are falsely presented | |
| | in order to drive the population to obedient behaviour, and to vaccination. | |
| | So please be critical, do your own research, and do not let the media manipulate you. | |

# 4 Findings

In summary, both studies provided insights into the linguistic features of fake news compared to real news. Fake news demonstrated more emotive language, assertiveness, and negativity in both social media text posts and TikTok transcribed speech. Additionally, various other linguistic traits, which were explored via other LIWC categories, played a role in distinguishing fake news from real news in both contexts.